\pdfoutput=1
\documentclass[11pt]{article}

\usepackage[]{acl}

\usepackage{times}
\usepackage{latexsym}

\usepackage[T1]{fontenc}
\usepackage[utf8]{inputenc}

\usepackage{microtype}

\usepackage{amsmath}
\usepackage{graphicx}
\usepackage{multirow}
\usepackage{booktabs}
\usepackage{subcaption}
\usepackage{tablefootnote}

\newcommand{\up}[1]{\textsuperscript{#1}}

\title{Multi-Task Learning for Depression Detection in Dialogs}

\author{Chuyuan Li\up{1}, Chloé Braud\up{2}, Maxime Amblard\up{1}\\
        \up{1} Universite de Lorraine, CNRS, Inria, LORIA, F-54000 Nancy, France  \\
        \up{2} IRIT, Université de Toulouse, CNRS, ANITI, Toulouse, France\\
        \up{1} \texttt{\{firstname.name\}@loria.fr}, 
        \up{2} \texttt{chloe.braud@irit.fr}
        }

\begin{document}
\maketitle
\begin{abstract}
Depression is a serious mental illness that impacts the way people communicate, especially through their emotions, and, allegedly, the way they interact with others.
This work examines depression signals in dialogs, a less studied setting that suffers from data sparsity.
We hypothesize that depression and emotion can inform each other, and we propose to explore the influence of dialog structure through topic and dialog act prediction.
We investigate a Multi-Task Learning (MTL) approach, where all tasks mentioned above are learned jointly with dialog-tailored hierarchical modeling.
We experiment on the DAIC and DailyDialog corpora -- both contain dialogs in English -- and show
important
improvements over state-of-the-art on depression detection (at best $70.6$\% F$_1$), which demonstrates the correlation of depression with  emotion and dialog organization and the power of MTL to leverage information from different sources.
\end{abstract}

%%%%%%%%%%%%%%%%%%%%%%%%%%%%%%%%%%%%%%%%%%%%%
\section{Introduction}
\label{sec:introduction}
Depression is a serious mental disorder that affects around $5$\% of adults worldwide.\footnote{\url{https://www.who.int/news-room/fact-sheets/detail/depression}} 
It comes with multiple causes and symptoms, leading to major disability, but is often hard to diagnose, with about half the cases not detected by primary care physicians \cite{cepoiu2008recognition}.
Automated detection of depression, sometimes associated to other mental health disorders, has been the topic of several studies recently, with a particular focus on social media data and online forums \cite{coppersmith2015clpsych,benton2017multitask, guntuku2017detecting, yates-etal-2017-depression,song-etal-2018-feature,akhtar2019all, rissola2021survey}.
The ultimate goal of such system would be to complement expert assessments, but such empirical studies are also valuable to better understand how communication is affected by health disorders. 
In this paper,
we propose to investigate depression detection within dialogs, 
a scenario less studied but more similar to the interviews with clinicians,
which allegedly involves dialog features and allows to also examine
how interaction is affected. 

However, depression detection suffers from data sparsity.
In fact, using social media data was a way to tackle this issue, including considering data generated by self diagnosed users, a method that leads to potentially noisy data and comes with ethical issues~\cite{chancellor2019taxonomy}.
We rather examine a dataset of 189 clinical interviews, the DAIC-WOZ \cite{gratch2014distress}, collected by experts to support the diagnosis of distress conditions. Participants are identified as depressive or not, and if so they receive a severity score.
A line of work proposed to overcome data scarcity by leveraging varied modalities, e.g. using audio as in \citet{alhanai2018detecting}. 
Previous approaches solely based on textual information relied on hierarchical contextual attention networks on word and sentence-level representations \cite{mallol2019hierarchical}, or multi-task learning (MTL) but limited to combing identification and severity prediction~\cite{qureshi2019multitask,dinkel2019text}, possibly with emotion~\cite{qureshi2020improving}.

Inspired by the latter approaches, we also propose to rely on 
MTL framework to help our model leveraging information from different sources.
We exploit three auxiliary tasks: emotion
-- 
naturally tied to mental health states --, but also dialog act and topic classification, hoping this shallow information about the dialog structure could further enhance the performance. 
Our architecture is classic, based on hard-parameter sharing~\cite{ruder2017overview}, 
simpler than the shared-private architecture in \cite{qureshi2020improving} but shown effective.
In order to take into account dialog organization, 
we advocate for a dialog-tailored hierarchical architecture with some tasks performed at the speech turn level and others at the document level. 

Our contributions are: 
(i) An empirical study on depression detection in dialogs, leveraging the power of multi-task learning to deal with data sparsity;
(ii) An extension of previous work in examining the effects of depression on dialog structure via shallow markers, i.e., dialog acts and topics, as a first step;
(iii) State-of-the-art results on depression detection in DAIC test set with $70.6$\% in F$_1$ at best.

%%%%%%%%%%%%%%%%%%%%%%%%%%%%%%%%%%%%%%%%%%%%%
\section{Related work}
\label{sec:related}

Within multi-task learning (MTL), a model has to learn shared representations to generalize the target task better. 
It improves the performance over single-task learning (STL) by leveraging commonalities or correlations between tasks.
Recent years have witnessed a series of successful applications in various NLP tasks, as in  
\citet{collobert2008unified, sogaard2016deep, ruder2017overview, ruder2019latent}, which
demonstrates the effectiveness of MTL
in learning
information from different but related sources. 
It also
tackles the data sparsity issue and reduces the risk of overfitting
\cite{mishra-etal-2017-learning,benton2017multitask,bingel2017identifying}.

\citet{joshi2019does} demonstrated the benefit of MTL for specific pairs of close health prediction tasks on tweets.
\citet{benton2017multitask} used MTL on social media data and achieved important improvements in predicting
several mental health signals, including suicide risks, depression, and anxiety, together with gender prediction.
With a focus on depression detection, the shared task AVEC in 2016 \cite{valstar2016avec} has brought out a series of multi-modal studies using vocal and visual features on the DAIC-WOZ dataset \cite{gratch2014distress}.
Some of which also explore text-level features: 
\citet{williamson2016detecting} used Gaussian Staircase Model with semantic content features and reported a SOTA score on the validation set.
\citet{alhanai2018detecting} and \citet{haque2018measuring} learned sentence embeddings with an LSTM network. However, their results on textual features are lower than SOTA by a large margin.
\citet{dinkel2019text} compared different word and sentence embeddings and various pooling strategies. Their best model is mean pooling with ELMo embeddings.
\citet{qureshi2019multitask, qureshi2020improving}
proposed MTL approaches 
in adding emotion intensity and depression severity (i.e., a regression problem) prediction to the main classification task.
They, however, found that the emotion-unaware model obtained the best result.
They used a monologue corpus for the emotion task,
a domain bias that possibly harms the performance.
On the contrary, we hypothesize that emotional information would benefit depression detection.
\citet{mallol2019hierarchical} used a hierarchical contextual attention network with static word embeddings within a single-task setting and then combined representations at the word and sentence levels. They reported at best $63$\% in F$_1$.
Recently, \citet{xezonaki2020affective} presented even better results, $70$\% in F$_1$, by augmenting the attention network with a conditioning mechanism based on effective external lexicons and incorporating the summary associated with each interview. 
We instead rely on MTL in this work, where incorporating external sources is more direct. 

None of the previous studies investigated potential links between depression and dialog structure.
We note that \citet{cerisara2018multi} explored MTL with sentiment\footnote{Sentiment and emotion are closely related with different function and/or granularity, cf. \citet{munezero2014sentemo}. \citet{cerisara2018multi} use three labels for sentiment: \textit{positive}, \textit{negative}, \textit{neutral}. In this paper, we use seven emotional labels: \textit{anger}, \textit{disgust}, \textit{fear}, \textit{happiness}, \textit{sadness}, \textit{surprise}, \textit{neutral}.} and dialog act prediction on Mastodon (a Twitter-like dataset), where both annotations are available, and found a positive correlation. 
To the best of our knowledge, we are the first to tackle depression detection in dialog transcriptions with the MTL approach and explore joint learning techniques with tasks related to the dialog structure.

%%%%%%%%%%%%%%%%%%%%%%%%%%%%%%%%%%%%
\section{Model Architecture}
\label{sec:model}
One condition generally assumed for success within MTL, at least in NLP, is that the primary and auxiliary tasks should be related \cite{ruder2017overview}.
The emotion-related task is thus a natural choice since it is linked to mental states. 
We hypothesize that depressive disorder can also affect how people interact with others
during conversations. 
We thus take a first step toward linking dialog structure and depression by examining shallow signals: dialog acts and topics.
In addition, since the information comes at different levels, we propose hierarchical modeling, from speech turns to documents. 

\paragraph{Baseline Model:} 
Our basic model is a two-level 
recurrent network, similar to the one in \citet{cerisara2018multi}. 
The input words are mapped to 
vectors using
word embeddings from scratch. The first level (\textit{turn}-level) takes the embeddings into a bi-LSTM network to obtain one vector for each turn. The second level (\textit{dialog}-level) takes a sequence of turns into an RNN network, and the output is finally passed into a linear layer for depression prediction. 

\begin{figure}
    \centering
    \includegraphics[width=.95\columnwidth]{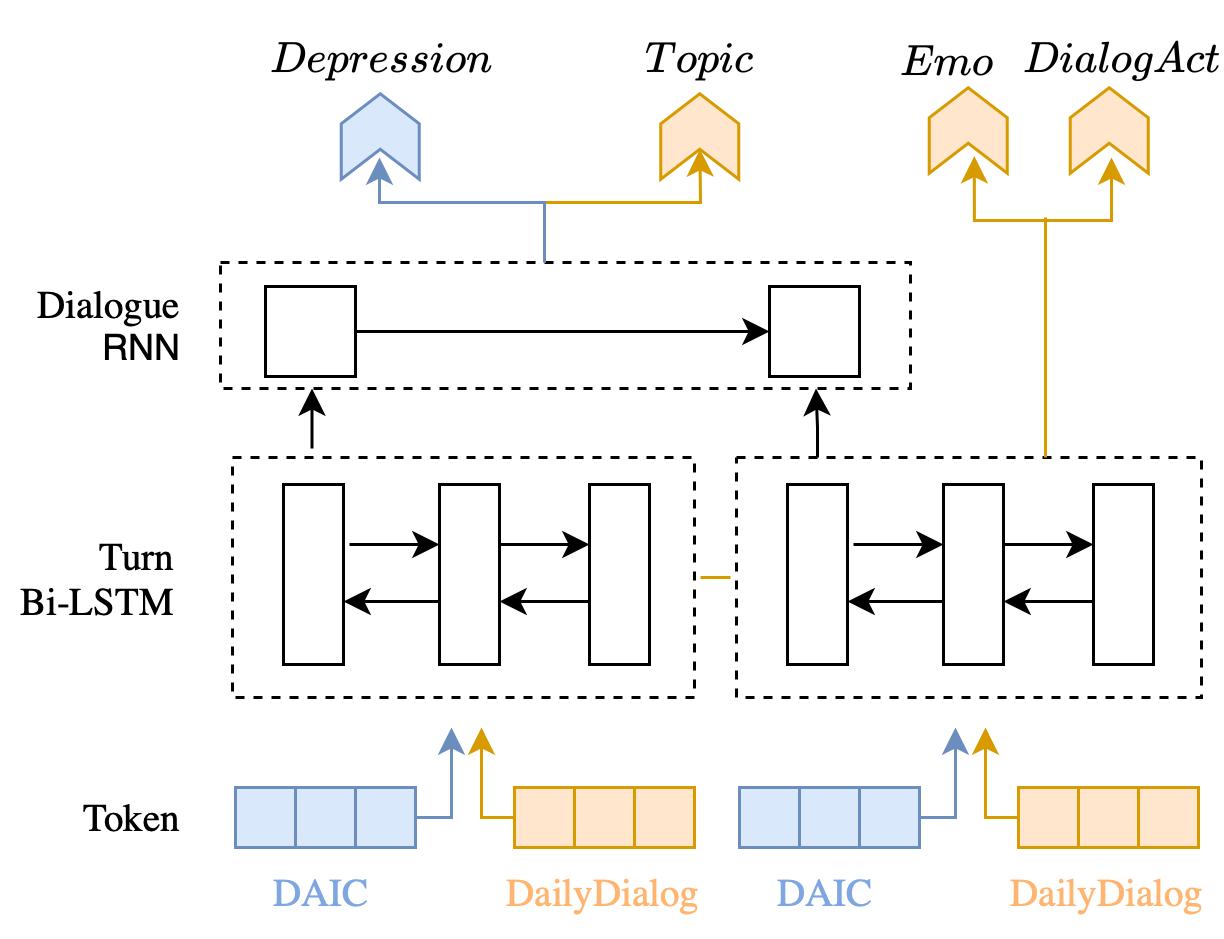}
    \caption{Multi-task fully shared hierarchical structure. Light blue is for DAIC dataset and depression task; orange is for DailyDialog and three auxiliary tasks.}
    \label{fig:structure}
\end{figure}

\paragraph{MTL Model:} The MTL architecture is composed of shared hidden layers and task-specific output layers (see Fig.~\ref{fig:structure}) and corresponds to the hard parameter sharing approach 
\cite{caruana1993multitask, caruana1997multitask, ruder2017overview}.
Since some auxiliary tasks are at the speech-turn level (i.e., emotion, dialog act) while others are annotated
at the document level 
(i.e., depression, topic), our architecture is hierarchical
and arranges task-specific output layers (MLP) at two levels. 
Speech-turn level emotion and dialog act information can be learned in the \textit{turn}-level LSTM network and transferred upwards to help depression and topic prediction. 
On the other hand, higher-level information can be backpropagated to update the network at the lower level.
The loss is simply
the sum of the losses for each task.
When it comes to the MTL setting, we set equal weight for each task as the standard choice. 

%%%%%%%%%%%%%%%%%%%%%%%%%%%%%%%%%%%%%%%%%%%%%
\section{Datasets}
\label{sec:datasets}

\paragraph{DAIC-WOZ:} This dataset 
is a subset of the DAIC corpus \cite{gratch2014distress}.\footnote{\url{https://dcapswoz.ict.usc.edu}}
It contains $189$ sessions (one session is one dialog with avg. $250$ speech turns) of two-party interviews between participants and Ellie -- an animated virtual interviewer controlled by two humans.
Table~\ref{tab:daic-subsets-origin} gives the 
partition of train ($107$), development ($35$), and test ($47$) sets.
Originally, patients are associated with a score related to the
Patient Health Questionnaire (PHQ-9): a patient is considered depressive if 
PHQ-9 $\geq10$ 
\cite{kroenke2002phq}.

\begin{table}[]
    \centering
    \begin{tabular}{llll}
        \toprule
        & Train & Dev & Test  \\
        \midrule
        Depressed & 77 & 23& 33 \\
        Non Depressed & 30 & 12 & 14 \\
        \midrule
        Total & 107 & 35 & 47 \\
        \bottomrule
    \end{tabular} %}
    \caption{Number of sessions (dialogs) in DAIC-WOZ.}
    \label{tab:daic-subsets-origin}
\end{table}

\paragraph{DailyDialog:}
This dataset \cite{li2017dailydialog}
contains $13,118$
two-party
dialogs (with averaged $7.9$ speech turns per dialog) for English learners,\footnote{\url{http://yanran.li/dailydialog}}
covering various topics from ordinary life to finance.
Three expert-annotated information are provided: $7$ emotions \cite{ekman1999emotions}, $4$ coarse-grain dialog acts,
and $10$ topics.
We select this corpus due to its large size, two-level annotations and high quality. 
The train set contains $>87k$ turns for emotions and dialog acts and $>11k$ dialogs for topics.
Detailed statistics are given in 
Appendix~\ref{append:dd-detail}.

\section{Experimental setup}
\label{sec:exp-setup}
\paragraph{Baselines: }
We compare our MTL results with: (1) Majority class where the model predicts all positive;
(2) Baseline single-task model (see Sec.~\ref{sec:model}); (3) State-of-the-art results on test set reported by \citet{mallol2019hierarchical} and \citet{xezonaki2020affective}.
We do not compare to \cite{williamson2016detecting, haque2018measuring, alhanai2018detecting, dinkel2019text, qureshi2020improving} who only report on the development set.

\paragraph{Evaluation Metrics: }
For depression classification
we follow \citet{dinkel2019text}
and report accuracy, macro-F$_1$, precision, and recall. For emotion analysis, we follow \citet{cerisara2018multi} and report macro-F$_1$.

\paragraph{Implementation Details: }
We implement our model with AllenNLP library \cite{gardner2018allennlp}.
We use the original separation of train, validation, and test sets for both corpora. 

The model is trained for a maximum of $100$ epochs with early stopping.
For STL as well as for MTL scenario, we optimize on macro-F$_1$ metric for depression classification. We use cross-entropy loss. 
The batch size is $4$ for DailyDialog and $1$ for DAIC (within the limit of GPU VRAM).
We use the tokenizer from spaCy Library \cite{honnibal2020spacy} and construct the word embeddings
by default with a dimension of $128$.
The \textit{turn} level has one hidden layer and $128$ output neurons. 
We tune \textit{document} RNN layers in $\{1, 2, 3\}$ and hidden size in $\{128, 256, 512\}$.
Model parameters are optimized using Adam \cite{kingma2014adam} with $1e-3$ learning rate.
Dropout rate is set to $0.1$ for both \textit{turn} and \textit{document} encoders. 
The source code is available at \url{https://github.com/chuyuanli/MTL4Depr}.

%%%%%%%%%%%%%%%%%%%%%%%%%%%%%%%%%%%%%%%%%%%%%
\section{Results and Discussion}
\label{sec:results}
\subsection{Depression Detection Results on DAIC}
\label{subsec:binary-class-daic}
Results using our MTL hierarchical structure are shown in Table~\ref{tab:daic-binary-result}, which are compared to SOTA models (at the top). 
Our baseline model is a single-task naive hierarchical model which obtains similar results (F$_1$ $44$) as the baseline model (NHN) in \citet{mallol2019hierarchical} (F$_1$ $45$).

Using the multi-task architecture, we get improvements when adding each task separately. We see more than a +$11.5$\% increase in F$_1$ when adding emotion (`+Emo') or topic (`+Top') classification task and, at best, +$16.9$\% with dialog acts (`+Diag'). This demonstrates the relevance of each task to the primary problem of depression detection, especially the interest of dialog acts. 
When adding topics, we observe a small drop in accuracy compared to STL while the F$_1$ is better, meaning that the prediction for minority class (non-depressive) improves.
Interestingly, in terms of accuracy, the tasks at different levels (depression `+Emo' and depression `+Diag')
seem to help more. We deduce that they help build a better local representation (speech turns) before
the global representation.

When jointly learning all four tasks -- combining depression detection with three auxiliary tasks (`+Emo+Diag+Top') --, all metrics improve. 
We obtain our best system with an improvement of +$26.7$\% in F$_1$ compared to STL baseline, outperforming the state-of-the-art with a +$7.6$\% increase compared to the best system in \citet{mallol2019hierarchical} and about +$0.5$\% compared to \citet{xezonaki2020affective}.
Depressed people tend to express specific emotions; it is thus natural to think that emotion is beneficial for the main task. 
These results indicate 
that both emotion and dialog structure help as they provide complementary information,
paving the way for new research directions with more fine-grained modeling of dialog structure for tasks in conversational scenarios.

\begin{table}[]
    \centering
    \resizebox{\columnwidth}{!}{
    \begin{tabular}{l rrrr}
    \toprule
    & F$_1$ & Prec. & Rec. & Acc. \\
    \midrule
    BSL Majority vote  & $41.3$ & $35.1$ & $50.0$ & $70.2$ \\
    \midrule
    \multicolumn{2}{l}{\textit{State-of-the-art}}\\
    NHN\tablefootnote{Naive hierarchical network (baseline).
    } \cite{mallol2019hierarchical}  & $45$ &- & $50$ &-\\
    HCAN\tablefootnote{Hierarchical contextual attention network.
    } \cite{mallol2019hierarchical}  & $63$ & - & $66$ & -\\
    HAN+L\tablefootnote{Hierarchical attention network with LIWC lexicon.} \cite{xezonaki2020affective}  & $70$ &- & $70$ &-\\
    \midrule
    \textit{Ours}\\
    STL Depression  & $43.9$ & $44.5$& $47.5$  & $63.8$\\
    MTL \hspace{0.1 in}+Emo  & $55.5$ & $56.2$& $61.6$  & $70.2$\\
    MTL \hspace{0.1 in}+Top & $55.6$ & $55.9$ & $56.8$ & $59.6$\\
    MTL \hspace{0.1 in}+Diag & $60.8$ & $60.6$ & $61.4$ & $66.0$ \\
    MTL \hspace{0.1 in}+Emo+Diag+Top & $\textbf{70.6}^\ast$ & $\textbf{70.1}$ & $\textbf{71.5}^\ast$ & $\textbf{74.5}$\\
    \bottomrule
    \end{tabular}}
    \caption{Depression detection results on DAIC.
    STL: single-task using DAIC only;
    MTL: multi-task using DAIC and
    adding classification for Emotion (+Emo), Topic (+Top), Dialog Act (+Diag) from DailyDialog. *Significantly better than SOTA performance with p-value $<0.05$.
    }
    \label{tab:daic-binary-result}
\end{table}

%%%%%%%%%%%%%%%%%%%%%%%%%%%%%%%%%%%%
\subsection{Analysis}
\label{sec:analysis}

\paragraph{Performance on Auxiliary Tasks: }
To better understand our model, we look at the performance of 
emotion, dialog act, and topic auxiliary tasks.
Directly comparing the results of our MTL approach (`+Emo+Diag+Top') with a STL architecture for each task, however,
seems
unfair. The optimized objective and structural complexity are different: the former is optimized on the depression detection task on two levels, while the latter is tuned on the target auxiliary task with either speech turn (emotion and dialog act) or full dialog (topic).
Unsurprisingly, the results show that the MTL system underperforms the basic STL structure for dialog acts and topics, with at best $67.8$ in F$1$ (MTL) \textit{vs.} $68.8$ (STL) for dialog acts, and $52.0$ (MTL) \textit{vs.} $52.4$ (STL) for topic classification.

For emotion, on the other hand, our best MTL system obtains $40.0$ in F$_1$ compared to $38.3$ for the STL baseline, showing the mutual benefit of both tasks.
Even though the score is lower than the SOTA for emotion classification ($51.0$ F$_1$ in \citet{qin2021co})\footnote{Precision: in \citet{qin2021co} authors report results on sentiment classification. It is yet unclear how they convert emotion annotation ($7$ labels) to sentiment ($3$ labels).}, we believe that refining our model for this task could lead to further improvements in depression detection.
In addition, we observe that our MTL approach is particularly beneficial for negative and rare emotion classes, with \textit{anger}, \textit{disgust} and \textit{sadness} gaining resp. $5\%$, $6\%$ and $1\%$ in F$_1$.
Finally, we conduct a manual inspection of the types of 
utterances (mostly questions) from Ellie, and classify them into high-level dialog acts: \textit{Backchannel, Comment, Opening, Other, Question}.\footnote{\textit{Backchannel} refers to phatic expressions such as \textit{yeah, hum mm}. Here we use different dialog acts from those in DailyDialog.}
We find that around $13$\% of the utterances are emotion-related, 
for instance ``things which make you mad / you feel guilty about, last time feel really happy'', etc.,
and that mentions of topics related to happiness or regret appear in almost all the interviews.
Dialog act distribution is shown in Table~\ref{tab:daic_da_propor}. We release our annotation to the community for future studies.

\begin{table}[]
    \centering
    \resizebox{\columnwidth}{!}{
    \begin{tabular}{lrrrrr}
    \toprule
    High-level DA & \# & \% & Sub-cat. & \# & \% \\
    \midrule
    \multirow{2}{*}{Question} & \multirow{2}{*}{$7,907$} & \multirow{2}{*}{$53\%$} &  Emo & $1,054$ & $13\%$ \\
    & & & Non-emo & $6,853$ & $87\%$ \\
    Backchannel & $3,231$ & $22\%$ & - & - & -\\
    Comment & $3,074$ & $20\%$ & - & - & -\\
    Opening & $611$ & $4\%$ & - & - & -\\
    Other& $171$ & $1\%$ & - & - & -\\
    \bottomrule
    \end{tabular}}
    \caption{High-level dialog act distribution of Ellie in DAIC-WOZ. \# and \% represent the number and percentage of Ellie's utterances, respectively.}
    \label{tab:daic_da_propor}
\end{table}

\paragraph{Effectiveness of Hierarchical Structure: }
To examine the effectiveness of hierarchical structure, we conduct ablation studies on the full multi-learning setting (`+Emo+Diag+Top'). For dialog RNN level, we use topic information; for turn level, we test either emotion or dialog act. The results are shown in Table~\ref{tab:albation-study}.
Unsurprisingly, both ablated models (`+Emo+Top' and `+Diag+Top') underperform the full model, with F$_1$ scores decreasing $\approx$ $6\%$ each. 
Without dialog act, all metrics drop, showing the importance of this information for dialog structure. 
Without emotion, recall drops dramatically while accuracy and precision increase, indicating that the model `+Diag+Top' predicts more positive classes but fails in negative ones, which could result in too many false positives in real-life scenarios.
On the other hand, when comparing hierarchical models (`+Emo+Top', `+Diag+Top', `+Emo+Diag+Top') with single-level models (`+Emo', `+Top', `+Diag'), we see considerable improvements in all metrics, and this holds for all auxiliary tasks. 
We can thus confirm the advantage of hierarchical structure for model performance.

\begin{table}[]
    \centering
    \resizebox{\columnwidth}{!}{
    \begin{tabular}{l rrrr}
    \toprule
    & F$_1$ & Prec. & Rec. & Acc. \\
    \midrule
    MTL \hspace{0.2 in}+Emo+Diag+Top & $\textbf{70.6}$ & $70.1$ & $\textbf{71.5}$ & $74.5$\\
    MTL \hspace{0.2 in}+Emo+Top & $64.4$ & $64.4$ & $64.4$ & $70.2$ \\
    MTL \hspace{0.2 in}+Diag+Top & $63.7$ & $\textbf{78.1}$ & $62.8$ & $\textbf{76.6}$ \\
    \bottomrule
    \end{tabular}}
    \caption{Ablation study on hierarchical structure.}
    \label{tab:albation-study}
\end{table}

%%%%%%%%%%%%%%%%%%%%%%%%%%%%%%%%%%%%
\section{Conclusion}
\label{sec:conclusion}
In this paper, we demonstrate the correlation between depression and emotion and show the relevance of features related to dialog structures via shallow markers: dialog acts and topics.
In the near future, 
we intend to investigate more refined modeling of dialog structures, possibly relying on discourse parsing~\cite{shi2019deep}. 
We would also like to explore depression severity classification as an extension to binary classification, possibly through a cascading structure: first detect depression and then classify the severity.
We intend to refine our work and report on cross-validation splits of the data to test the stability of the model, an issue even more crucial when dealing with sparse data with possibly representativeness problem.
A further
step will be to investigate the generalization of our model
to other mental health disorders.

%%%%%%%%%%%%%%%%%%%%%%%%%%%%%%%%%%%%%%%%%%%%%
\section*{Acknowledgement}
The authors thank the anonymous reviewers for their insightful comments and suggestions. This work was supported by the PIA project ``Lorraine Université d’Excellence'', ANR-15-IDEX-04-LUE, as well as the CPER LCHN (Contrat de Plan État-Région - Langues, Connaissances et Humanités Numériques).
It was also partially supported by the ANR (ANR-19-PI3A-0004) through the AI Interdisciplinary Institute, ANITI, as a part of France's ``Investing for the Future --- PIA3'' program, and through the project AnDiAMO (ANR-21-CE23-0020).
Experiments presented were carried out in secured clusters on the Grid'5000 testbed. We would like to thank the Grid'5000 community (\url{https://www.grid5000.fr}).

%%%%%%%%%%%%%%%%%%%%%%%%%%%%%%%%%%%%%%%%%%%%%
\section*{Ethical Considerations}
The goal of such systems is not to replace human healthcare providers. All these systems may be used only in support to human decision. The principle of leaving the decision to the machine would imply major risks for decision making in the health field, a mistake that in high-stakes healthcare settings could prove detrimental or even dangerous.

Another issue is the representativeness of the data. Currently, it is very complex to access patients in order to have more examples. The institutional complexity leads researchers to systematically use the same data set, creating a bias between the representation of the pathology, in particular for mental ones whose expression can take very varied forms. This also implies defining a variation in relation to a normative use of language that comes with a strong risk in this type of approach.

Moreover, we carefully select the dialog corpora used in this paper to control for potential biases and personal information leakage.
We only work with interview transcription, with no audio or visual information. For the text part, all the participant's name have been marked out with pseudo-ID.

%%%%%%%%%%%%%%%%%%%%%%%%%%%%%%%%%%%%%%%%%%%%%
\bibliography{biblio}
\bibliographystyle{acl_natbib}

%%%%%%%%%%%%%%%%%%%%%%%%%%%%%%%%%%%%%%%%%%%%%
\clearpage
\newpage
\appendix
\section{Auxiliary Tasks Class Distribution in DailyDialog}
\label{append:dd-detail}
Table~\ref{tab:daily-emotion}, Table~\ref{tab:daily-da}, and Table~\ref{tab:daily-topic} show the number and percentage of emotion, dialog act, topic for each subset, resp.

\begin{table}[h!]
    \centering
    \resizebox{\columnwidth}{!}{
    \begin{tabular}{lrr rr rr}
        \toprule
        \multirow{2}{*}{Emotion} & \multicolumn{2}{c}{Train} & \multicolumn{2}{c}{Dev} & \multicolumn{2}{c}{Test}\\
        \cmidrule(lr){2-3} \cmidrule(lr){4-5} \cmidrule(lr){6-7}
        & \# & \% & \# & \% & \# & \% \\
        \midrule
        0-no emotion & $72,143$ & $82.8$ & $7,108$ & $88.1$ & $6,321$ & $81.7$\\
        1-anger & $827$ & $0.9$ & $77$ & $1.0$ & $118$ & $1.5$\\
        2-disgust & $303$ & $0.3$ & $3$ & $0.04$ & $47$ & $0.6$\\
        3-fear & $146$ & $0.2$ & $11$ & $0.1$ & $17$ & $0.2$\\
        4-happiness & $11,182$ & $12.8$ & $684$ & $8.5$ & $1019$ & $13.2$\\
        5-sadness & $969$ & $1.1$ & $79$ & $1.0$ & $102$ & $1.3$\\
        6-surprise & $1,600$ & $1.8$ & $107$ & $1.3$ & $116$ & $1.5$\\
        \midrule
        Utt. Total & $87,170$ & $100.0$ & $8,069$ & $100.0$ & $7,740$ & $100.0$ \\
        \bottomrule
    \end{tabular}}
    \caption{Emotion distribution in train, dev. and test sets.}
    \label{tab:daily-emotion}
\end{table}

\begin{table}[h!]
    \centering
    \resizebox{\columnwidth}{!}{
    \begin{tabular}{lrr rr rr}
        \toprule
        \multirow{2}{*}{Dialog Act} & \multicolumn{2}{c}{Train} & \multicolumn{2}{c}{Dev} & \multicolumn{2}{c}{Test}\\
        \cmidrule(lr){2-3} \cmidrule(lr){4-5} \cmidrule(lr){6-7}
        & \# & \% & \# & \% & \# & \% \\
        \midrule
        1-inform & $39,873$ & $45.7$ & $3,125$ & $38.7$ & $3,534$ & $45.7$\\
        2-question & $24,974$ & $28.6$ & $2,244$ & $27.8$ & $2,210$ & $28.6$\\
        3-directive & $12,242$ & $16.3$ & $1,775$ & $22.0$ & $1,278$ & $16.5$\\
        4-commissive & $8,081$ & $9.23$ & $925$ & $11.5$ & $718$ & $9.3$\\
        \midrule
        Utt. Total & $87,170$ & $100.0$ & $8,069$ & $100.0$ & $7,740$ & $100.0$ \\
        \bottomrule
    \end{tabular}}
    \caption{Dialog act distribution in train, dev. and test sets.}
    \label{tab:daily-da}
\end{table}

\begin{table}[h!]
    \centering
    \resizebox{\columnwidth}{!}{
    \begin{tabular}{lrr rr rr}
        \toprule
        \multirow{2}{*}{Topic} & \multicolumn{2}{c}{Train} & \multicolumn{2}{c}{Dev} & \multicolumn{2}{c}{Test}\\
        \cmidrule(lr){2-3} \cmidrule(lr){4-5} \cmidrule(lr){6-7}
        & \# & \% & \# & \% & \# & \% \\
        \midrule
        1-ordinary life & $2,975$ & $26.8$ & $418$ & $41.8$ & $252$ & $25.2$\\
        2-school life & $453$ & $4.1$ & $0$ & $0$ & $34$ & $3.4$\\
        3-culture \& education & $50$ & $0$ & $0$ & $0.0$ & $5$ & $0.5$\\
        4-attitude \& emotion & $616$ & $5.5$ & $1$ & $0.0$ & $50$ & $0.5$\\
        5-relationship & $3,879$ & $34.9$ & $129$ & $12.9$ & $384$ & $38.4$\\
        6-tourism & $860$ & $7.7$ & $124$ & $12.4$ & $79$ & $7.9$\\
        7-health & $205$ & $1.8$ & $41$ & $4.1$ & $21$ & $2.1$\\
        8-work & $1,574$ & $14.2$ & $215$ & $21.5$ & $135$ & $1.4$\\
        9-politics & $105$ & $0.9$ & $13$ & $1.3$ & $13$ & $1.3$\\
        10-finance & $399$ & $3.6$ & $59$ & $5.9$ & $27$ & $2.7$\\
        \midrule
        Total & $11,118$ & $100.0$ & $1,000$ & $100.0$ & $1,000$ & $100.0$ \\
        \bottomrule
    \end{tabular}}
    \caption{Topic distribution in train, dev. and test sets.}
    \label{tab:daily-topic}
\end{table}

\end{document}